\documentclass[conference,a4paper]{APSIPA2021}
\usepackage{amsmath}
\usepackage{graphicx}
\usepackage{multirow}
\usepackage{booktabs}
\usepackage{threeparttable}
\usepackage{makecell}
\usepackage{amssymb}
\usepackage{float}
\usepackage{url}
\usepackage{pifont}
\usepackage{subfigure}
\usepackage{tikz}
\usepackage{etoolbox}

\usepackage[backend=biber,style=ieee,]{biblatex}
\usepackage{geometry}
\usepackage{fancyhdr}

\fancypagestyle{firststyle}{
  \fancyhf{}
  \fancyhead[C]{2023 Asia Pacific Signal and Information Processing Association Annual Summit and Conference (APSIPA ASC)}
}
\begin{document}
\title{AI-Generated Image Detection using a  Cross-Attention Enhanced  Dual-Stream Network}

\author{
\authorblockN{
Ziyi Xi,
Wenmin Huang,
Kangkang Wei,
Weiqi Luo and
Peijia Zheng
}

\authorblockA{
GuangDong Province Key Laboratory of Information Security Technology,   Sun Yat-sen University, GuangZhou, China 
 \\
E-mail: luoweiqi@mail.sysu.edu.cn \quad
xiziyi@mail2.sysu.edu.cn}
}

\maketitle

\thispagestyle{firststyle}

\begin{abstract}
  With the rapid evolution of AI Generated Content (AIGC), forged images produced through this technology are inherently more deceptive and require less human intervention compared to traditional Computer-generated Graphics (CG). However, owing to the disparities between CG and AIGC, conventional CG detection methods tend to be inadequate in identifying AIGC-produced images. To address this issue, our research concentrates on the text-to-image generation process in AIGC. Initially, we first  assemble two text-to-image databases utilizing two distinct AI systems, DALL·E2 and DreamStudio.  Aiming to holistically capture the inherent anomalies produced by AIGC, we  develope a robust dual-stream network comprised of a residual stream and a content stream. The former employs the Spatial Rich Model (SRM) to meticulously extract various texture information from images, while the latter seeks to capture additional forged traces in low frequency, thereby extracting complementary information that the residual stream may overlook. To enhance the information exchange between these two streams, we  incorporate a cross multi-head attention mechanism. Numerous comparative experiments are performed on both databases, and the results show that our detection method consistently outperforms traditional CG detection techniques across a range of image resolutions. Moreover, our method exhibits superior performance through a series of robustness tests and cross-database experiments. When applied to widely recognized traditional CG benchmarks such as SPL2018 and DsTok, our approach significantly  exceeds the capabilities of other existing methods in the field of CG detection. 
\end{abstract}

\section{Introduction}
\begin{figure}[t]
\centering
\subfigure[DALL$\cdot$E2]{
\begin{minipage}[b]{0.47\linewidth}
    \includegraphics[width=0.9\textwidth]{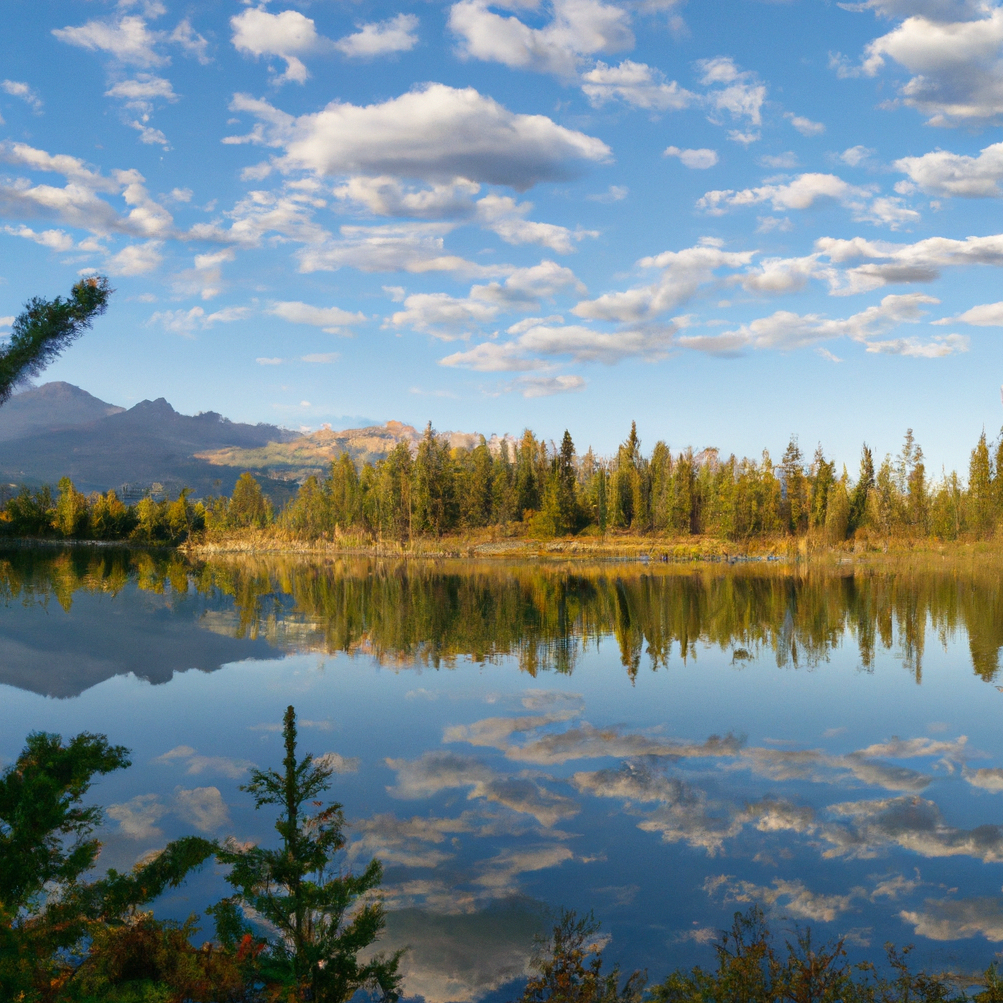}\vspace{3pt}
    \includegraphics[width=0.9\textwidth]{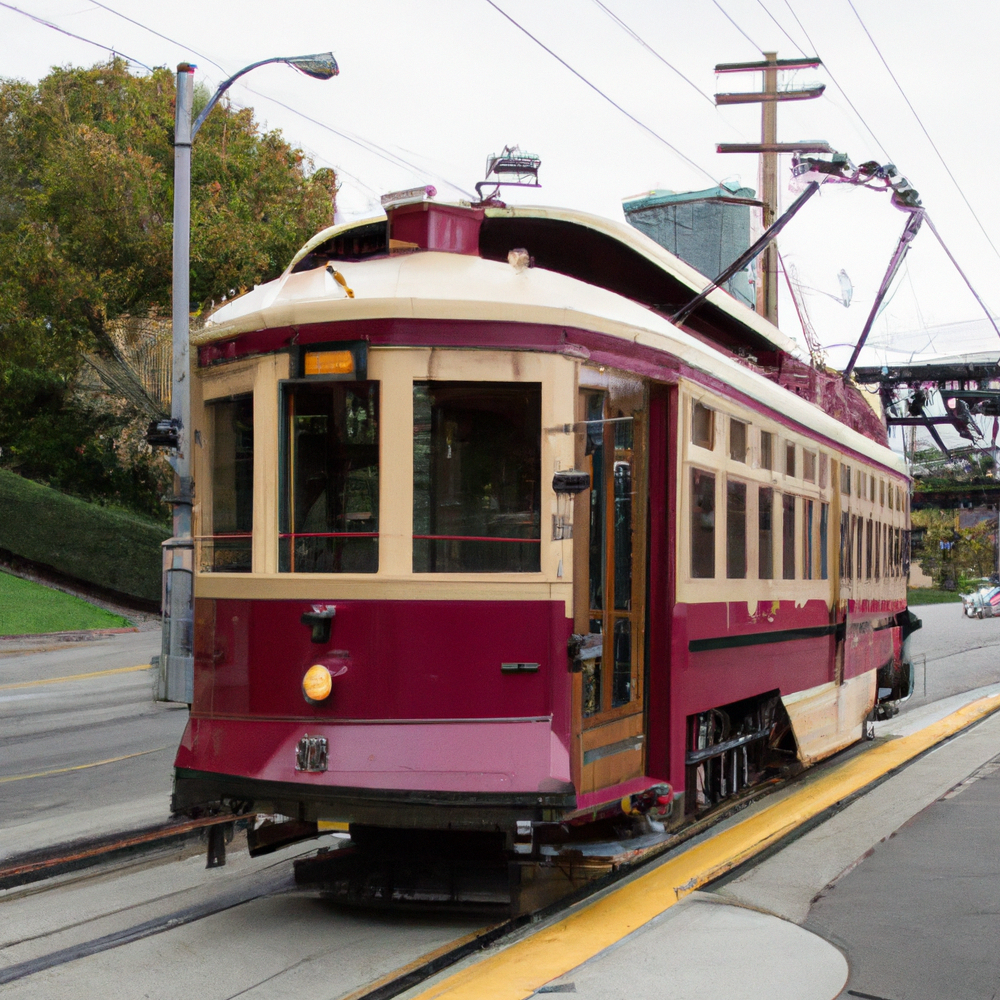}
\end{minipage}}
\subfigure[DreamStudio]{
\begin{minipage}[b]{0.47\linewidth}
    \includegraphics[width=0.9\textwidth]{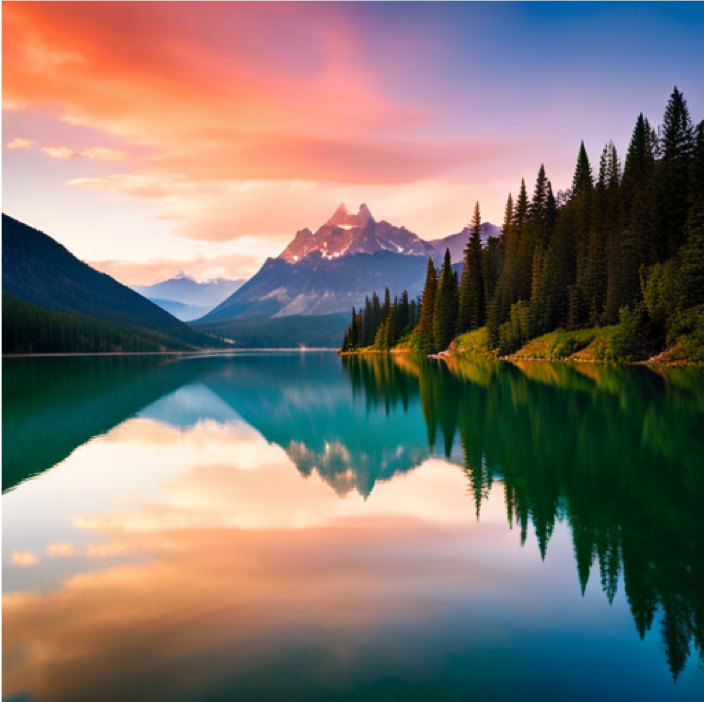}\vspace{3pt}
    \includegraphics[width=0.9\textwidth]{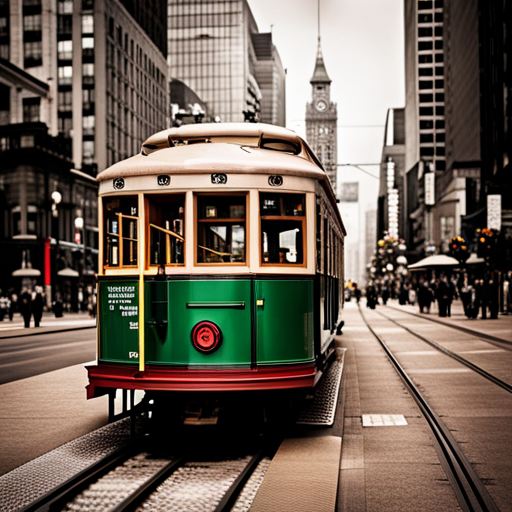}
\end{minipage}}
\caption{Generated images utilizing two different AI systems, DALL$\cdot$E2 and DreamStudio, respectively. The prompt for the first row is:``The lake is like a mirror, reflecting the trees and mountains surrounding it." The prompt for the second row is: ``The charming old-fashioned streetcars that trundled along the tracks added a touch of nostalgia to the scene." }
\label{fig_example}
\end{figure}

AI-generated content (AIGC), which pertains to the production or creation of content through artificial intelligence systems, has exerted a significant influence across various application scenarios, particularly with the emergence of large-scale models. The content produced by generative AI models comprises text, image, audio, video and cross-modal transformations like text-to-image, text-to-audio and more \cite{cao2023comprehensive}.  Among these, text-to-image has garnered enormous attention in social media platforms, primarily owing to its potential to supplant human visual design.  In this field, several representative models have emerged, one such model is DALL$ \cdot$E, which is trained using Variational Autoencoders, another important model is VQGAN-CLIP based on Generative Adversarial Networks (GANs).  Nowadays the diffusion model has become the core approach in text-to-image generation, with popular models like Stable Diffusion, Disco Diffusion, MidJourney, and DALL$\cdot$E2 etc \cite{zhang2023text}. By typing the prompt containing expected image concepts, attributes and styles, users can cultivate an realistic or artistic image which is significantly correlative with the prompt.  Some image examples are shown in Fig.\ref{fig_example}.  Compared to conventional Computer-generated  Graphics (CG) technology, more ease of operation, low capitalized cost and high-quality image feedback have propelled text-to-image  generation into the forefront of internet. Therefore how to distinguish text-to-image graphics (denoted as T2I for short) from Photographs (PG) generated by a digital camera has emerged as a pressing issue that requires resolution.  \par

To our knowledge, only a limited number of works have focused on distinguishing between T2I and PG.   Certain methods capable of distinguishing between these two image types capitalize on familiar classification techniques prevalent in computer vision, such as  ResNet \cite{he2016deep, sha2022fake} and EfficientNet \cite{efficientnet}. However, these methods often fall short in providing a comprehensive analysis of inherent artifacts in T2I images and usually necessitate pre-training. As an alternative, we could harness contemporary methods established for distinguishing between  CG  and PG, given that these methods are adept at detecting variations in lighting, color, texture, and other attributes present in PG. To date, several conventional methods have been proposed for CG and PG detection. For instance, Quan et al. \cite{quan2018} employed a $7\times7$ convolutional kernel in the initial layer to adaptively extract noise information. Yao et al. \cite{yao2018} designed three fixed high-pass filtering kernels to capture high-frequency information from images. Zhang et al. \cite{Zhang2020} developed a module comprising plain convolutional structures to extract correlations between adjacent pixels in RGB color space. He et al. \cite{He2020} applied a Gaussian filter as a preprocessing step on the input image to achieve a robust network. Quan et al. \cite{quan2020} leveraged the  SRM  \cite{srm} to extract residual information from different channels. Other prevalent solutions include strategies based on transfer learning \cite{ref12,ref13,ref14,Yao2022}, the integration of Convolutional Neural Network (CNN) and Recurrent Neural Networks (RNN) \cite{SPL2018}, and those based on self-supervised learning \cite{selfsupervised}, among others.  \par
While the methods mentioned above prove effective for CG  and  PG  detection, directly applying these techniques to T2I and PG  detection might not yield optimal results. This limitation is due to the unique generative mechanisms present in T2I and CG. Inspired by successful approaches in CG detection, such as those presented in \cite{He2020, quan2020}, we propose a novel cross-attention enhanced dual-stream network specifically designed for T2I and PG detection.
Our method incorporates two streams: a residual stream and a content stream. The residual stream employs a  SRM \cite{srm} residual extraction module, which meticulously extracts texture information from the images. Conversely, the content stream focuses on the low-frequency aspects of the images, utilizing a dedicated network structure. The information extracted from both streams is downscaled using  CNN modules, and then fused via a cross multi-head attention mechanism situated in the network's middle layer. After independent learning, the features from both streams are combined through channel concatenation, and a classifier is subsequently utilized to discern whether the image is T2I or PG. To ascertain the efficacy of our proposed method, we  generate  two   T2I databases. Comprehensive comparative and ablation experiments have affirmed the superiority of our method. In summary, our contributions in this paper include:
\begin{itemize}
\item We propose a novel dual-stream framework that combines a residual stream and a content stream to comprehensively explore the generation traces of  T2I. Moreover, we introduce a cross multi-head attention mechanism to enhance information exchange between the two streams.
\item To assess the performance of our approach, we  construct  two T2I databases using two AI systems: DALL$\cdot$E2 and DreamStudio. Each database consists of 20,000 generated images covering various scenarios.
\item Our method excels in detection performance on our custom databases (DALL$\cdot$E2 and DreamStudio) and outperforms comparable methods on established CG detection datasets, specifically DsTok and SPL2018. Moreover, we have conducted several ablation experiments to validate the effectiveness of our model design.
\end{itemize}
\vspace{2mm}

\begin{figure}[ht]
\centering
\subfigure[Input Image]{
     \includegraphics[width=0.16\textwidth]{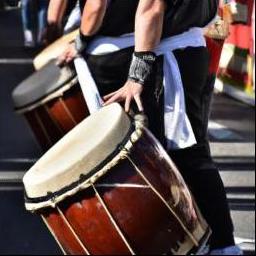}} 
 \subfigure[1st Order: Spam 14h]{
     \includegraphics[width=0.16\textwidth]{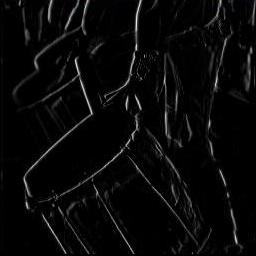}} 
 \subfigure[2nd Order: Spam 12h]{
     \includegraphics[width=0.16\textwidth]{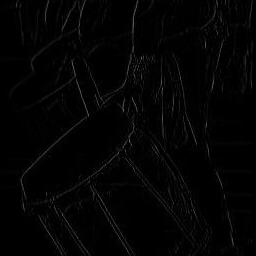}} 
 \subfigure[EDGE $3\times3$: Spam 14h]{
     \includegraphics[width=0.16\textwidth]{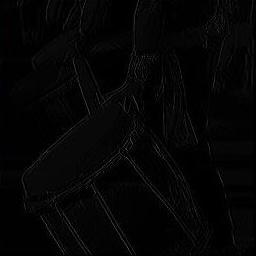}} 
 \caption{An example image and its corresponding residual information obtained using three different high-pass filters.}
  \label{fig:srm}
\end{figure}

\section{Proposed Method}

\begin{figure*}[t!]
\centering
\subfigure[The network architecture of the  proposed model]{
\centering
\includegraphics[width=160mm]{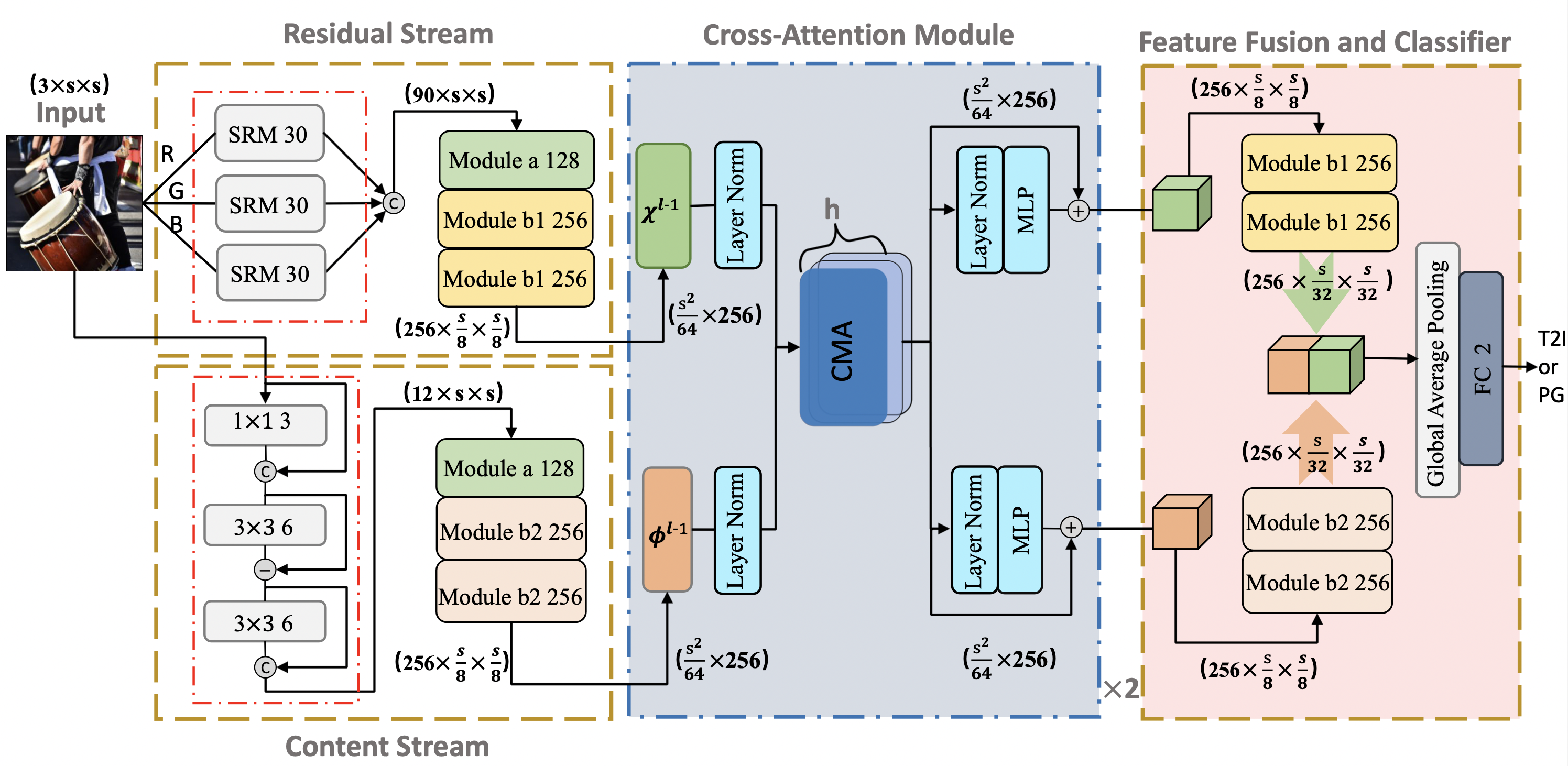}
}
\vspace{-3mm}
\subfigure[Module a]{
 \begin{minipage}[b]{0.19\linewidth}
    \includegraphics[width=0.8\textwidth]{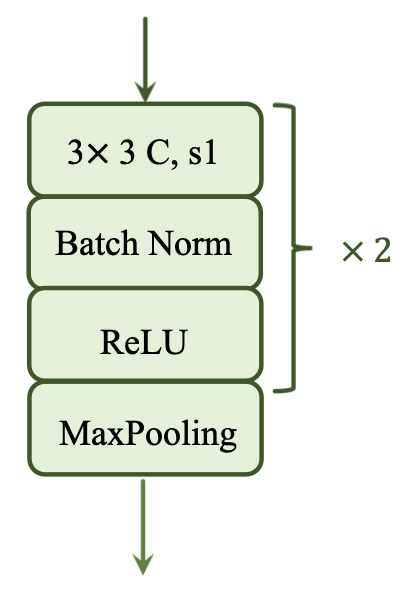}\vspace{2pt}
\end{minipage}}
\subfigure[Module b1]{
\begin{minipage}[b]{0.23\linewidth}
    \includegraphics[width=0.8\textwidth]{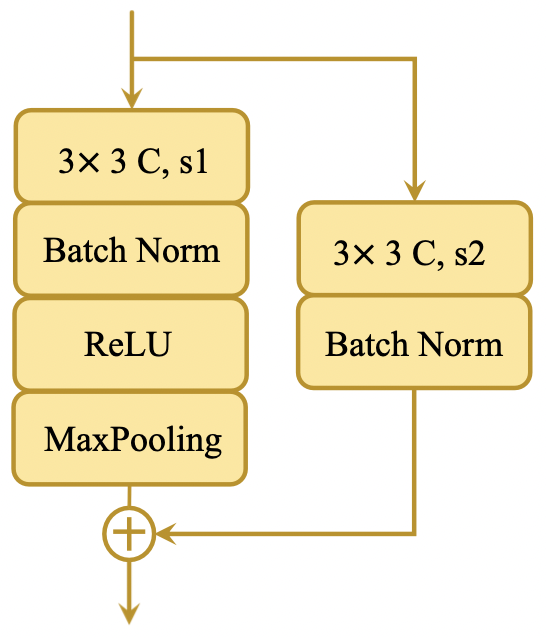}\vspace{2pt}
\end{minipage}}
\hspace{-3mm}
\subfigure[Module b2]{
\begin{minipage}[b]{0.128\linewidth}
    \includegraphics[width=0.8\textwidth]{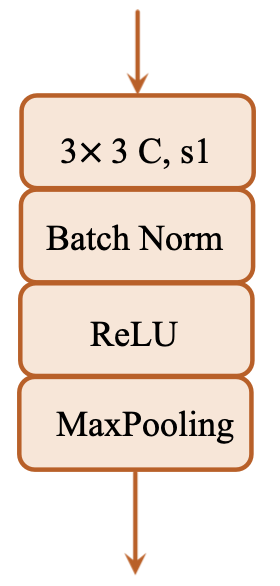}\vspace{2pt}
\end{minipage}}
\hspace{-3mm}
\subfigure[MLP]{
\begin{minipage}[b]{0.13\linewidth}
    \includegraphics[width=0.8\textwidth]{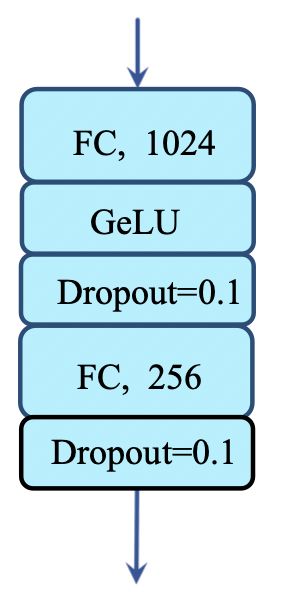}\vspace{2pt}
\end{minipage}}
\caption{The network architecture of the proposed model, and four modules used in the proposed model. The variable $h$ corresponds to the number of headers. FC means fully connected layer.  $C$ represents the number of output channels, s1 means the stride of convolution kernel is 1.The symbol $\bigoplus$ denotes element-wise summation, and $\large \ominus$ means element-wise subtraction, \textcircled{c} means channel-wise concatenation. }
\label{fig:architecture}
\vspace{-3mm}
\end{figure*}

The network architecture of the proposed model is visualized in Fig. \ref{fig:architecture}. In the following subsections, we will provide a detailed description of the residual stream, content stream, cross-attention module, feature fusion and classifier utilized in the proposed model, respectively. Additionally, we will demonstrate the loss function employed in our model. 

\subsection{Residual Stream} 
 
Previous studies in CG detection  \cite{quan2020, yao2018} and image steganalysis \cite{deng2019fast, wei2022universal} have shown that forgery artifacts are often present in the high-frequency components of images. To better detect these artifacts, several studies have employed multiple high-frequency filters from the SRM  during image preprocessing to generate corresponding residual images. In our proposed model, we adopt a similar approach inspired by the methodologies presented in papers \cite{wei2022universal, quan2020}. As shown in the Fig. 3, the dimension of the input image is $3\times s \times s$,  where 3 means the color channels and $s \times s$ represents the resolution of input image.  We begin by applying 30 SRM kernels individually to each of the RGB color channels in the input images. In this way, the input image is transformed into the residual domain for further feature analysis. Fig. \ref{fig:srm} illustrates an example of the resulting residual images after filtering with three typical SRM kernels. This approach generates a total of 90 different residual images, aiming to capture a comprehensive representation of the multi-channel residual information. To facilitate further learning, we downscale 90 residual images using a custom-designed stacked CNN modules, which consist of one Module a and two Module b1.   Module a and Module b1 are defined as follows: 

\begin{itemize}
\item Module a: As shown in Fig. \ref{fig:architecture}-(b),  it  comprises two consecutive groups of $3\times3$ convolutional layer, batch normalization (BN), ReLU, finally followed by a maximum pooling layer of size $3\times3$ with a stride of 2.
\item Module b1:  As shown in Fig. \ref{fig:architecture}-(c), it includes  two parts: convolutional downsampling and pooling downsampling. The pooling downsampling involves a sequence of $3\times3$ convolutional layer, BN, and a ReLU, followed by a maximum pooling layer of size $3\times3$ with a stride of 2. In parallel, the input feature maps are downscaled in another branch using a $3\times3$ convolutional layer with a stride of 2. The downscaled feature maps from both branches are fused together through element-wise addition, then activated by ReLU function.
\end{itemize}

After 90 residual feature maps through one Module a and two Module b1 in sequence, we can obtain $ \chi^{l-1} \in R^{256 \times \frac{s}{8} \times \frac{s}{8}}$,  as labeled in Fig. 3.

\begin{figure}[ht]
\centering
\subfigure{
\begin{minipage}[b]{0.35\linewidth}
    \includegraphics[width=0.8\textwidth]{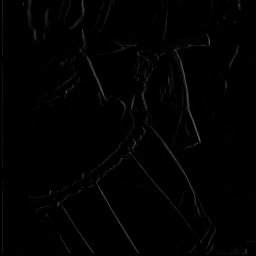}
\end{minipage}
\hspace{-5mm}
\begin{minipage}[b]{0.35\linewidth}
    \includegraphics[width=0.8\textwidth]{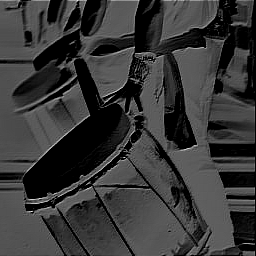}
\end{minipage}
\hspace{-5mm}
\begin{minipage}[b]{0.35\linewidth}
    \includegraphics[width=0.8\textwidth]{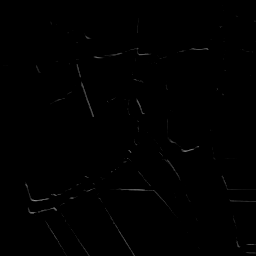}
\end{minipage}}
\hspace{-5mm}
\subfigure{
\begin{minipage}[b]{0.35\linewidth}
    \includegraphics[width=0.8\textwidth]{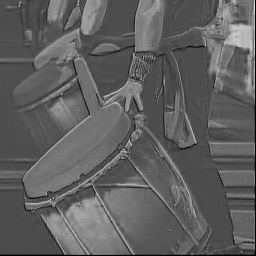}
\end{minipage}
\hspace{-5mm}
\begin{minipage}[b]{0.35\linewidth}
    \includegraphics[width=0.8\textwidth]{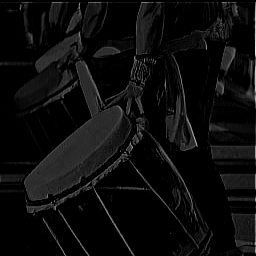}
\end{minipage}
\hspace{-5mm}
\begin{minipage}[b]{0.35\linewidth}
    \includegraphics[width=0.8\textwidth]{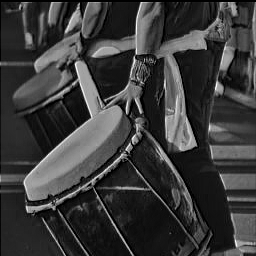}
\end{minipage}}
\caption{The visualization of $\Tilde{\ell}$ with 6 channels.}
\vspace{-7mm}
\end{figure}

\subsection{Content Stream}
As mentioned earlier, the residual stream primarily focuses on image texture while disregarding the underlying content information. Unlike traditional steganography, where the modifications are usually minimal and do not disrupt the mid-to-low frequency information of the original image, images generated by T2I or CG often exhibit significant statistical differences compared to images captured by a camera. Therefore, to address the limitations of relying solely on the residual stream, we propose a content stream that specifically targets forgery traces in the mid-to-low frequency range.
\par 
In the proposed content stream, we first utilize a $1\times 1$ convolution to generate three composite images which integrate the R, G, and B channels. This step is performed to enhance the breadth of channel information. Next, we concatenate the above three channels and the RGB image to obtain feature maps with 6 channels(denoted as $\ell$). Then, pass the $\ell$ through a $3\times3$ convolution, and $conv_{3\times3}(\ell)$ is subtracted from $\ell$ to obtain the feature maps $\Tilde{\ell}$, as Eq. (1). Subsequently, apply $3\times3$ convolution to $\Tilde{\ell}$ for further learning, and then {concatenate} the $\Tilde{\ell}$ and $Conv_{3\times 3}(\Tilde{\ell})$ along the channel dimension to obtain $\ell_{out}$ with 12 channels, as Eq. (2). Fig. 4 visualizes $\Tilde{\ell}$, where it can be observed that four out of the six channels exhibit distinct content variations in the image, while the remaining two channels primarily capture edge  information.  
 
\begin{align}
\centering
\Tilde{\ell} &= \ell - conv_{3\times3}(\ell)\\ 
\ell_{out} &= Concat(\Tilde{\ell}, conv_{3\times3}(\Tilde{\ell)}) 
\end{align}
 
Following that, $\ell_{out}$ are downscaled through the stacked CNN modules which involve one Module a and two Module b2. Finally get the feature map $\phi^{l-1} \in R^{256 \times \frac{s}{8} \times \frac{s}{8}}$.  As shown in Fig. \ref{fig:architecture}-(d),  the  Module b2 consists of $3\times3$ convolutional layer, BN layer, ReLU layer, and Maximum pooling layer of size $3 \times 3$ and a stride of 2.  

\subsection{Cross-Attention Module}
To facilitate the interactive fusion of dual-stream, we propose a cross-attention module to exchange information between dual-stream. The details are as follows: 

 First, flattening the output of two streams $\chi^{l-1}$ and $\phi^{l-1}$ respectively, and exchange the last two dimensions to obtain $\chi^{l-1}, \phi^{l-1} \in R^{\frac{s^2}{64} \times 256}$, followed by a layer normalization. Then, send them to the subsequent cross multi-head attention module (CMA), as shown in Fig. 5. Its principle and implementation details are as follows: 
 
The ViT \cite{vit} is a Transformer-based model for image classification. It treats the input image as a sequence of patches and maps each patch to a single vector using linear mapping. Self-attention (SA) mechanism, the core of transformer, utilizes three variables: Q (Query), K (Key), and V (Value). In essence, it computes the attention weights between Query token and Key token, and multiplies them with the corresponding Value token associated with each Key. The formula for  SA  can be expressed as follows, $d_k$ means the dimension of token:
\begin{align}
SA = softmax(\frac{QK^T}{\sqrt{d_k}})V
\end{align}
\par 
 Denotes the number of heads is $h$. We split $\chi^{l-1}, \phi^{l-1}$ into $h$ parts, attaining $\chi_i^{l-1},\phi_i^{l-1} \in R^{\frac{s^2}{64}\times \frac{256}{h}}$, where $i \in \{1,2,...,h\}.$ 
Different from SA, CMA calculates the degree of correlation between different vectors at each space position. In each head $i$, We can obtain the Query $Q_{\chi_i^{l-1}} \in R^{\frac{s^2}{64}\times \frac{256}{h}}$  and Key $K_{\phi_i^{l-1}}\in R^{\frac{s^2}{64}\times \frac{256}{h}}$ by linear projection respectively. Following that, ${\phi_i'}$ is obtained by multiplying the attention matrix with $\phi_i^{l-1}$. $d_k$ is $\frac{256}{h}$ here.\par
\begin{align}
\phi_{i}' &= softmax(\frac{Q_{\chi_i^{l-1}} \times K_{\phi _i^{l-1}}^{T}}{\sqrt{d_k}})  \phi_{i}^{l-1} 
\end{align}
\par
Subsequently, the $\phi_{i}'$ obtained from each head are concatenated together to obtain the $\phi'$ relevant to $\chi^{l-1}$, and add it to the $\chi^{l-1}$ to get the output $\chi^{l}$.  {Similarly, we can obtain $\phi^l$.} After that, a layer normalization and an MLP module are applied, as Fig. 3. The cross-attention module utilizes transformer encoder structure in general as paper \cite{vit} and repeats it for twice.  After through the cross-attention module, we can obtain the outputs $\in R^{\frac{s^2}{64} \times 256}$. 
\begin{align}
\chi^{l} &= \chi^{l-1} + Concat(\phi_1',\phi_2',....,\phi_h') 
\end{align}
 
\subsection{Feature Fusion and Classifier}
After through the cross-attention module, we need to convert the output from cross-attention module back to the feature maps with its original dimensions, swap the last two dimensions before reshaping and obtain the $\phi^{l}, \chi^{l} \in R^{C\times \frac{s}{8} \times \frac{s}{8} }$. The downsampling process further progresses as the feature maps are passed through a series of stacked CNN modules {which involve two Module b1/b2.} Afterwards, the feature maps are concatenated together by channel. Finally, classification is performed utilizing global average pooling and fully connected layer.

\begin{figure}[t!]
\begin{center}
\includegraphics[width=89mm]{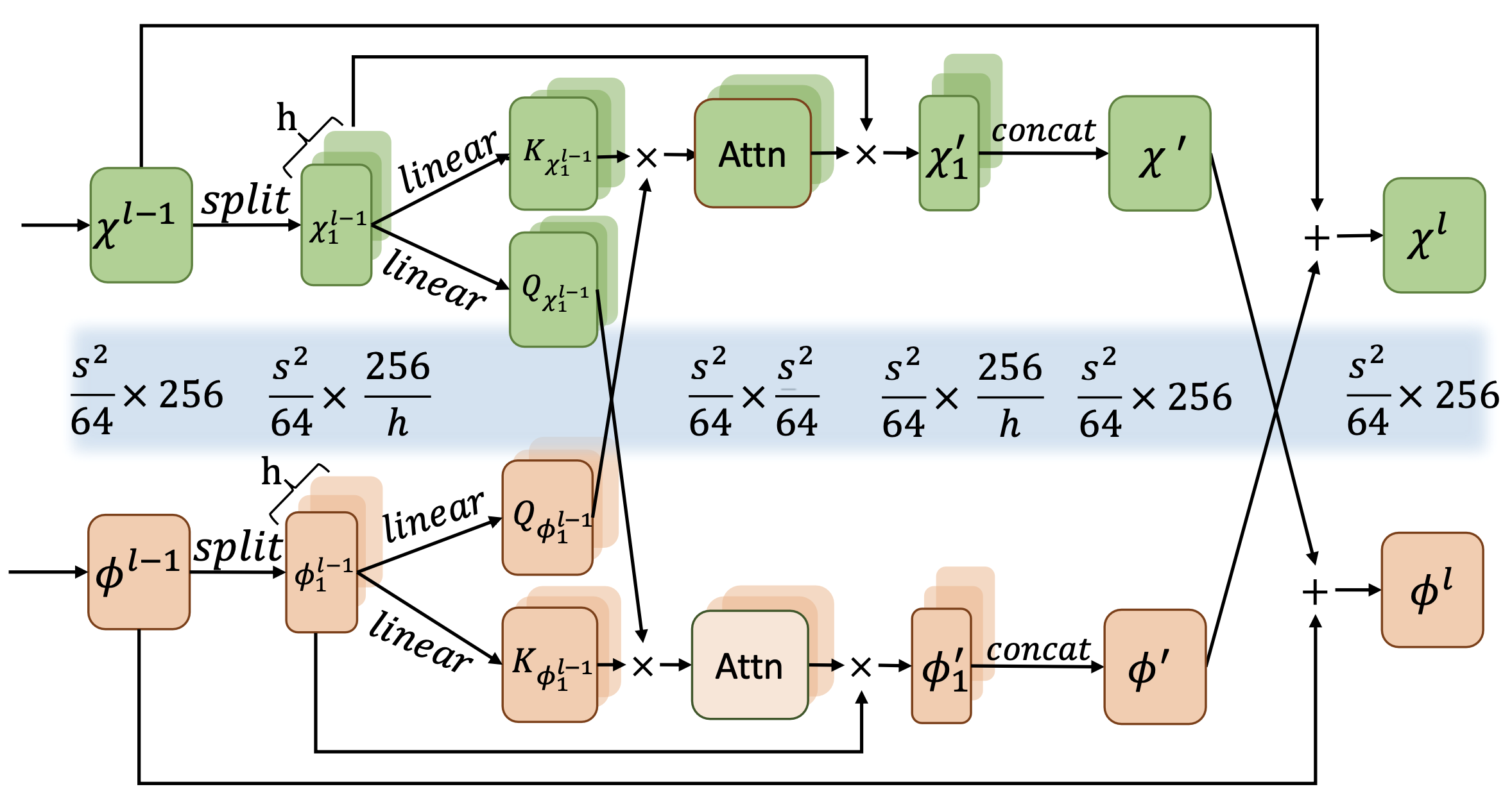}
\end{center}
\caption{The specific calculation process of the proposed CMA mechanism.}
\end{figure}

\subsection{Loss Function}
 
The  loss function used in the proposed model is the cross entropy loss, which is defined as follows. 
\begin{align}
\mathrm{Loss}=-\frac{1}{N} \sum_{i=1}^N\left[y_i \cdot \log p\left(y_i\right)+\left(1-y_i\right) \cdot \log \left(1-p\left(y_i\right)\right)\right]
\end{align}

Here, $N$ stands for the number of samples in a batch. The symbol $y_i$ denotes the true label of the i-th sample, where $y_i \in \{0,1\}$. On the other hand, $p(y_i)$ indicates the predicted probability associated with the true label $y_i$.

\begin{figure*}[t!]
\centering
\subfigure[Flowers]{
\rotatebox{90}{\scriptsize{~~~~~~DreamStudio~~~~~~~~~~~~~~DALL$\cdot$E2~~~~~~~~~~~~~~~~~ALASKA}}
\begin{minipage}[b]{0.13\linewidth}
\includegraphics[width=1\linewidth]{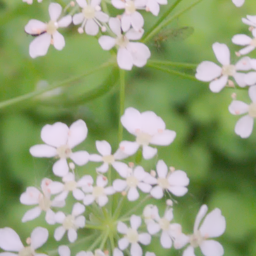} 
\includegraphics[width=1\linewidth]{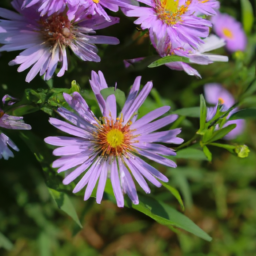} 
\includegraphics[width=1\linewidth]{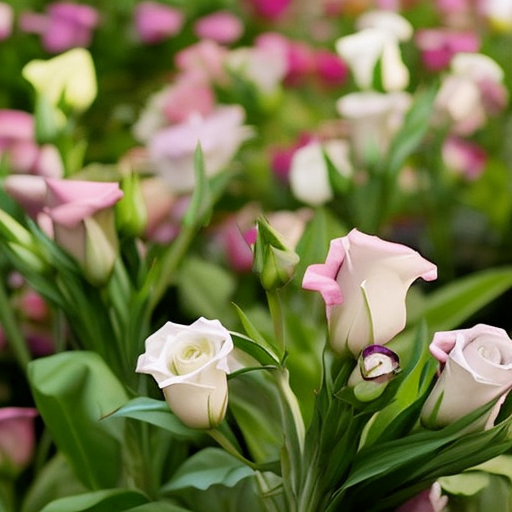}
\end{minipage}}
 \hspace{-2.5mm}
\subfigure[Animals]{
\begin{minipage}[b]{0.13\linewidth}
\includegraphics[width=1\linewidth]{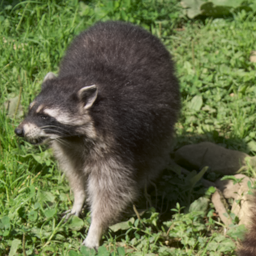} 
\includegraphics[width=1\linewidth]{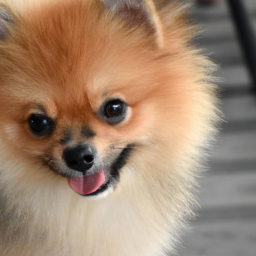} 
\includegraphics[width=1\linewidth]{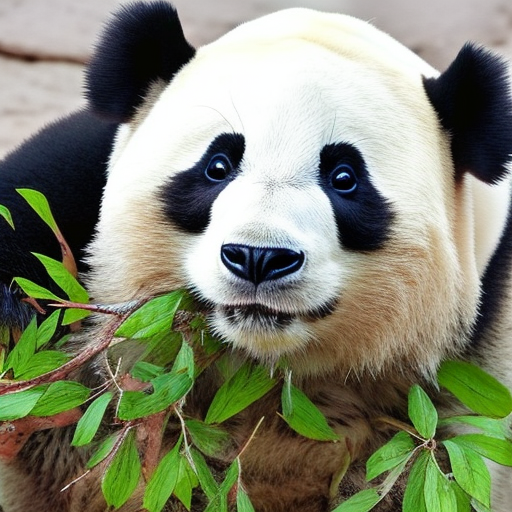}
\end{minipage}}
\hspace{-2.5mm}
\subfigure[Hill]{
\begin{minipage}[b]{0.13\linewidth}
\includegraphics[width=1\linewidth]{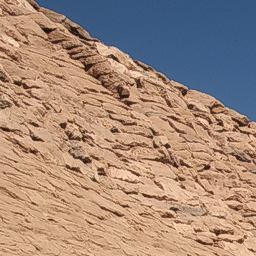} 
\includegraphics[width=1\linewidth]{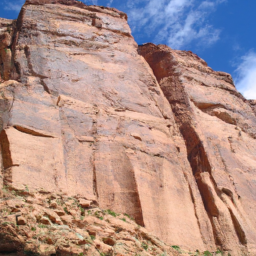} 
\includegraphics[width=1\linewidth]{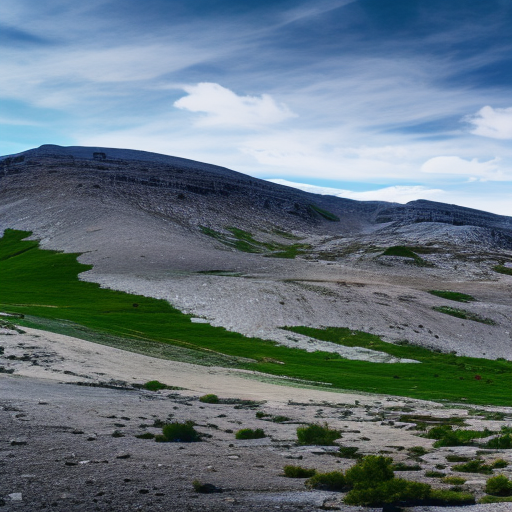}
\end{minipage}}
\hspace{-2.5mm}
\subfigure[Buildings]{
\begin{minipage}[b]{0.13\linewidth}
\includegraphics[width=1\linewidth]{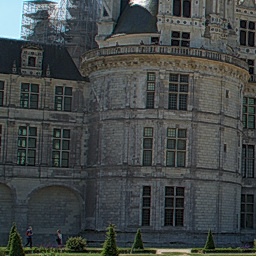} 
\includegraphics[width=1\linewidth]{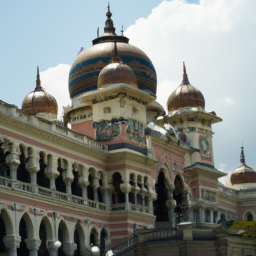} 
\includegraphics[width=1\linewidth]{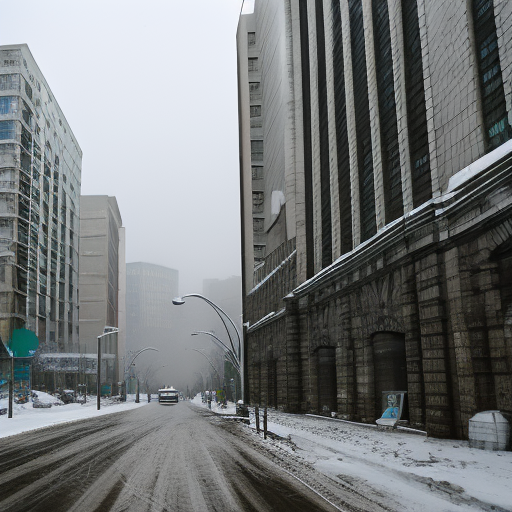}
\end{minipage}}
\hspace{-2.5mm}
\subfigure[Clouds]{
\begin{minipage}[b]{0.13\linewidth}
\includegraphics[width=1\linewidth]{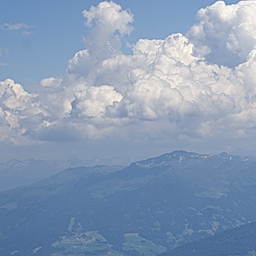} 
\includegraphics[width=1\linewidth]{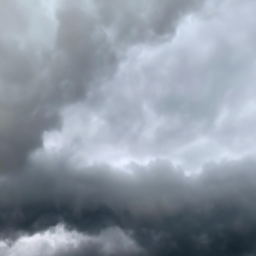} 
\includegraphics[width=1\linewidth]{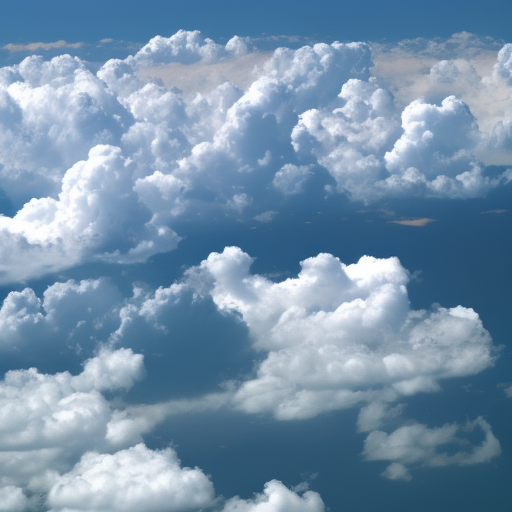}
\end{minipage}}
\hspace{-2.5mm}
\subfigure[Sculptures]{
\begin{minipage}[b]{0.13\linewidth}
\includegraphics[width=1\linewidth]{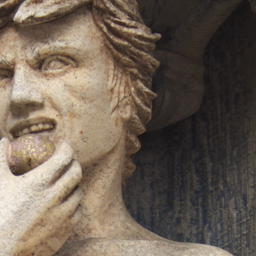} 
\includegraphics[width=1\linewidth]{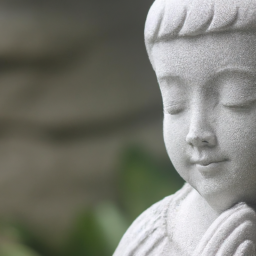} 
\includegraphics[width=1\linewidth]{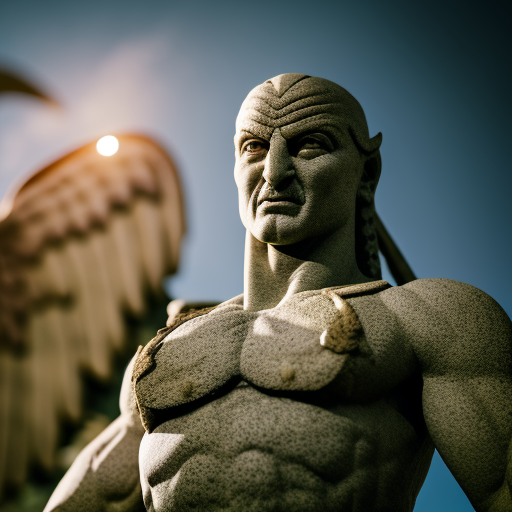}
\end{minipage}}
\hspace{-2.5mm}
\subfigure[Streets]{
\begin{minipage}[b]{0.13\linewidth}
\includegraphics[width=1\linewidth]{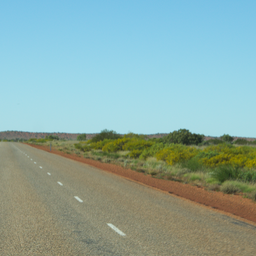} 
\includegraphics[width=1\linewidth]{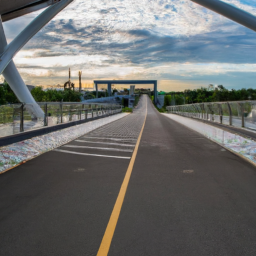} 
\includegraphics[width=1\linewidth]{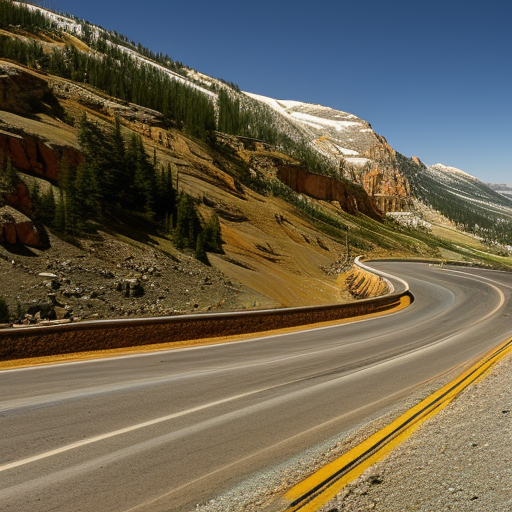}
\end{minipage}}
\hspace{-2.5mm}
\caption{Some image samples from ALASKA, DALL·E2, and DreamStudio under seven different scenarios.}
\label{fig:scenes}
\vspace{-5mm}
\end{figure*}

\section{Experiments result}
In this section, we begin by providing a detailed explanation of the process for creating two T2I databases, along with the implementation details of our model. Subsequently, we conduct comparative experiments on T2I databases, comparing our approach with seven contemporary CG detection methods  and ResNet18\cite{sha2022fake}. Additionally, we assess the robustness of our model against various post-processing operations and employ it for CG detection. Finally, we design ablation experiments to illustrate the validity of the proposed model.

\subsection{Constructing  Databases and Experimental Setup} 
Firstly, we randomly select 20,000 uncompressed PG images of size $256\times256$ from the ALASKA database\footnote{https://alaska.utt.fr/}, which is widely used in image steganalysis and forensics. To construct the T2I  dataset, we start by extracting specific spots or tangible objects from the selected PG images, such as sports fields, lakes, the sun, indoor scenes, Gothic architecture, and more. We then utilize ChatGPT to expand these keywords into 5,000 prompts. For each prompt, we use the APIs of two modern AI systems, namely DALL·E2\footnote{https://openai.com/product/dall-e-2} and DreamStudio\footnote{https://beta.dreamstudio.ai/generate}, to generate four different images. Finally, we create two databases, referred to as DALL·E2 and DreamStudio, each containing 20,000 PG images and their corresponding T2I images with similar scenes. Some image examples are shown in Fig. \ref{fig:scenes}. \par
It is important to note that the original sizes of the DALL·E2 and DreamStudio images are $256\times256$ and $512\times512$, respectively. To ensure uniformity, we resize all PG and T2I images to three different resolutions: $256\times256$, $128\times128$, and $64\times64$. Following the resizing, we apply JPEG compression to the images using a randomly selected quality factor ranging from 75 to 95. Consequently, we have a total of six distinct databases generated by the two AI systems, each incorporating the three resolutions. For each database, we partition the 20,000 PG-T2I image pairs into three mutually exclusive sets: 12,000 pairs for training, 3,000 pairs for validation, and the remaining 5,000 pairs for testing purposes.  

We employ an Adam optimizer with an initial learning rate of $2\times10^{-4}$ and a batch size of 64. The learning rate is reduced by a factor of 0.1 every 30 epochs, and the total number of training epochs is set to 120. The number of heads (i.e. $h$)  is set as 8. For training, we utilize 4 NVIDIA TITAN Xp GPUs. During evaluation, we consider the True Positive Rate (TPR) and the True Negative Rate (TNR) as our performance indicators.  Once this manuscript is accepted, we will provide the source code of our model online \footnote{https://github.com/zoie-ui/AI-Generated-Image-Detection}, allowing readers to replicate our experimental results. 

\begin{table*}[ht]
    \centering
    \caption{Comparative results for different methods on DALL$\cdot$E2 and DreamStudio at different resolutions. The highlighted values denoted the best result  in the corresponding case. The horizontal line means that the method has no output at this resolution.}
    \begin{tabular}{@{} *{10}{c}|*{9}{c} @{}}
      \Xhline{0.8pt}
      \multirow{3}*{Methods}&\multicolumn{9}{c}{DALL$\cdot$E2} \vline & \multicolumn{9}{c}{DreamStudio}\\
      \Xcline{2-10}{0.4pt} \Xcline{11-19}{0.4pt}
       & \multicolumn{3}{c}{256$\times$256} & \multicolumn{3}{c}{128$\times$128} & \multicolumn{3}{c}{64$\times$64}\vline  & \multicolumn{3}{c}{256$\times$256} & \multicolumn{3}{c}{128$\times$128} & \multicolumn{3}{c}{64$\times$64}\\
       \Xcline{2-10}{0.4pt} \Xcline{11-19}{0.4pt}
       & \scriptsize TPR & \scriptsize TNR & \scriptsize ACC & \scriptsize TPR & \scriptsize TNR & \scriptsize ACC & \scriptsize TPR & \scriptsize TNR & \scriptsize ACC & \scriptsize TPR & \scriptsize TNR & \scriptsize ACC & \scriptsize TPR & \scriptsize TNR & \scriptsize ACC & \scriptsize TPR & \scriptsize TNR & \scriptsize ACC\\
      \Xhline{0.4pt}
      ResNet18\cite{sha2022fake} & 96.3 & 94.8 & 95.6 & 94.0 & 92.3 & 93.2 & 87.4 & 85.8 & 86.6 & 97.8 & 96.3 & 97.1 & 97.5 & 97.0 & 97.2 & 93.3 & 93.7 & 93.5 \\
       Quan\cite{quan2018} &  97.6 & 97.4 & 97.5 & 96.8 & 95.9 & 96.3 & -- & -- & --& 98.2 & 98.6 & 98.4 & 98.5 & 97.3 & 97.9 & -- & -- & -- \\
      Yao\cite{yao2018}  &  96.3 & 94.5 & 95.3 & 96.3 & 94.2 & 95.3 & 86.1 & 88.3 & 87.2 & 97.1 & 97.5 & 97.3 & 97.6 & 97.0 & 97.3 & 91.3 & 92.3 & 91.8 \\
SPL2018\cite{SPL2018}    &  98.3 & 98.1 & 98.2 & 97.6 & 96.8 & 97.2 & 92.2 & 90.3 & 91.3  & 98.9 & 99.0 & 99.0  & 99.3 & 98.8 & 99.1 & 96.5 & 95.3 & 95.9\\
He\cite{He2020}         &  98.4 & 98.1 & 98.3 & 97.2 & 97.1 & 97.2 & 93.1 & 90.4 & 91.8 & 99.0 & 98.1  & 98.5 & 99.2 & 98.7 & 99.0 & 97.2 & 96.2 & 96.7\\
HcNet\cite{Zhang2020}      &  98.6 & 98.6 & 98.6 & 97.5 & 96.0 & 96.8 & 90.9 & 88.1 & 89.5 & 98.8 & 99.0 & 98.9 & 99.2 & 97.9 & 98.5 & 96.0 & 95.9 & 96.0\\    
QuanNet\cite{quan2020}       &  98.6 & 98.4 & 98.5 & 98.5 & 97.2 & 97.9 & \textbf{93.2} & 87.9 & 90.6 & 98.3 & 98.4 & 98.4 & 99.2 & 98.9 & 99.1 & 97.6 & 97.0 & 97.3\\   
CGNet\cite{Yao2022}      &  98.4 & 98.1 & 98.3 & 97.9 & \textbf{98.0} & 98.0 & 92.5 & \textbf{93.1} & 92.8 & 99.4 & 99.2 & 99.3 & 99.3 & 98.9 & 99.1 & \textbf{98.0} & 97.6 & \textbf{97.9} \\   
Ours       &  \textbf{99.3} & \textbf{99.1} & \textbf{99.2} & \textbf{98.6}& 97.9 & \textbf{98.3} & 93.1 & \textbf{93.1} & \textbf{93.1} & \textbf{99.5} & \textbf{99.6} & \textbf{99.5} & \textbf{99.4} & \textbf{99.5} & \textbf{99.5} & 97.7 & \textbf{97.8} & 97.8\\ 
    \Xhline{0.8pt}
    \end{tabular}
    \label{Tab:main}
\end{table*}

\begin{table*}[htb]
    \centering
    \caption{Detection accuracy for different post-processing on the DALL$\cdot$E2 test dataset at a resolution of $256\times 256$. The numbers in parentheses represent the ranking among all methods for the same post-processing.}
    \label{tab:post1}
    \begin{tabular}{*{9}{c}}
      \toprule
      DALL$\cdot$E2 & Chromaticity &  Brightness & Contrast & Sharpness & Rotation & Gaussian Blur & Mean Blur  & Average Acc \\
      \midrule
      ResNet18\cite{sha2022fake} & 94.0 \textcolor{blue}{(8)} & 91.3 \textcolor{blue}{(8)} & 82.3 \textcolor{blue}{(9)} & 94.8 \textcolor{blue}{(3)} & 53.4 \textcolor{blue}{(9)} & 86.6 \textcolor{blue}{(6)} & 71.5 \textcolor{blue}{(3)} &  82.0  \\
      Quan\cite{quan2018} & 96.7 \textcolor{blue}{(6)}& 94.5 \textcolor{blue}{(6)} & 92.7 \textcolor{blue}{ (4)} &  96.9 \textcolor{blue}{ (1)} &  63.1 \textcolor{blue}{ (6)} & 
      89.9 \textcolor{blue}{ (3)} &  60.4 \textcolor{blue}{ (8)} & 84.9
      \\
      Yao\cite{yao2018} & 92.1 \textcolor{blue}{ (9)} & 86.9 \textcolor{blue}{ (9)} & 86.9 \textcolor{blue}{ (8)} & 77.9 \textcolor{blue}{(9)} & 68.8 \textcolor{blue}{(2)} & 59.2 \textcolor{blue}{(9)} & 60.4 \textcolor{blue}{(8)} & 76.0
      \\
      SPL2018\cite{SPL2018} & 95.7 \textcolor{blue}{(7)} & 93.9 \textcolor{blue}{(7)} & 90.7 \textcolor{blue}{(6)} & 91.3 \textcolor{blue}{(7)} & 69.9 \textcolor{blue}{(1)} & 86.2 \textcolor{blue}{(7)} & 69.1 \textcolor{blue}{(5)} & 85.3
      \\
      He\cite{He2020} & 97.3 \textcolor{blue}{(4)} & 94.8 \textcolor{blue}{(4)} & 90.3 \textcolor{blue}{(7)} & 93.6 \textcolor{blue}{(5)} & 58.1 \textcolor{blue}{(8)} & 89.4 \textcolor{blue}{(4)} & 71.5 \textcolor{blue}{(3)}  & 84.9
      \\
      HcNet\cite{Zhang2020} & 97.5 \textcolor{blue}{(3)} & 94.8 \textcolor{blue}{(4)} & 91.3 \textcolor{blue}{(5)} & 93.9 \textcolor{blue}{(4)} & 64.1 \textcolor{blue}{(5)} & 92.7 \textcolor{blue}{(1)} & 85.1 \textcolor{blue}{(1)}  & \textbf{88.5}
      \\
      QuanNet\cite{quan2020} & 97.8 \textcolor{blue}{(1)} & 95.4 \textcolor{blue}{(1)} & 94.5 \textcolor{blue}{(2)} & 96.5 \textcolor{blue}{(2)} & 64.7 \textcolor{blue}{(4)} & 87.7 \textcolor{blue}{(5)} & 67.8 \textcolor{blue}{(6)} & 86.3
      \\
      CGNet\cite{Yao2022} & 97.2 \textcolor{blue}{(5)} & 95.4 \textcolor{blue}{(1)} & 94.0 \textcolor{blue}{(3)} & 86.4 \textcolor{blue}{(8)} & 61.5 \textcolor{blue}{(7)} & 81.3 \textcolor{blue}{(8)} & 67.7 \textcolor{blue}{(7)} & 83.4
      \\
      Ours & 97.7 \textcolor{blue}{(2)} & 95.2 \textcolor{blue}{(3)} & 95.7 \textcolor{blue}{(1)} & 91.7 \textcolor{blue}{(6)} & 67.4 \textcolor{blue}{(3)} & 90.9 \textcolor{blue}{(2)} & 80.6 \textcolor{blue}{(2)}  & \textbf{88.5}
      \\
\bottomrule
\end{tabular}
\end{table*}

\begin{table*}[htb]
    \centering
    \caption{Detection accuracy for different post-processing on the DreamStudio test dataset at a resolution of $256\times 256$. The numbers in parentheses represent the ranking among all methods for the same post-processing.        } 
     \label{tab:post2}
    \begin{tabular}{*{10}{c}}
      \toprule
      DreamStudio & Chromaticity &  Brightness & Contrast & Sharpness & Rotation & Gaussian Blur & Mean Blur & Average Acc \\
      \midrule
        ResNet18\cite{sha2022fake} & 95.9 \textcolor{blue}{(9)} & 91.7 \textcolor{blue}{(7)} & 85.3 \textcolor{blue}{(9)} & 89.5 \textcolor{blue}{(3)} & 56.2 \textcolor{blue}{(9)} & 67.1 \textcolor{blue}{(1)} & 66.7 \textcolor{blue}{(1)} &  78.9  \\
      Quan\cite{quan2018} & 97.5 \textcolor{blue}{(5)}& 92.4 \textcolor{blue}{(6)} & 92.7 \textcolor{blue}{(6)} &  88.8 \textcolor{blue}{(4)} &  78.8 \textcolor{blue}{(2)} & 
      57.2 \textcolor{blue}{(8)} &  57.8  \textcolor{blue}{(8)} & 80.7
      \\
      Yao\cite{yao2018} & 96.3 \textcolor{blue}{(8)} & 94.9 \textcolor{blue}{(2)} & 93.2 \textcolor{blue}{(3)} & 83.5 \textcolor{blue}{(9)} & 72.9 \textcolor{blue}{(4)} & 50.2 \textcolor{blue}{(9)} & 50.5 \textcolor{blue}{(9)} & 77.4
      \\
      SPL2018\cite{SPL2018} & 97.1 \textcolor{blue}{(7)} & 92.9 \textcolor{blue}{(5)} & 91.3 \textcolor{blue}{(7)} & 85.9 \textcolor{blue}{(7)} & 85.0 \textcolor{blue}{(1)} & 57.4 \textcolor{blue}{(7)} & 59.7 \textcolor{blue}{(5)} & 81.3
      \\
      He\cite{He2020} & 97.3 \textcolor{blue}{(6)} & 89.4 \textcolor{blue}{(8)} & 90.2 \textcolor{blue}{(8)} & 83.9  \textcolor{blue}{(8)} & 63.1 \textcolor{blue}{(6)} & 60.3 \textcolor{blue}{(3)} & 61.8 \textcolor{blue}{(2)} & 78.0
      \\
      HcNet\cite{Zhang2020} & 98.2 \textcolor{blue}{(3)} & 94.7 \textcolor{blue}{(3)} & 92.8 \textcolor{blue}{(5)} & 87.8 \textcolor{blue}{(6)} & 62.4 \textcolor{blue}{(7)} & 57.9 \textcolor{blue}{(6)} & 58.3 \textcolor{blue}{(7)} & 78.9
      \\
      QuanNet\cite{quan2020} & 97.9 \textcolor{blue}{(4)} & 89.1 \textcolor{blue}{(9)} & 93.1 \textcolor{blue}{(4)} & 88.3 \textcolor{blue}{(5)} & 58.5 \textcolor{blue}{(8)} & 59.7 \textcolor{blue}{(4)} & 61.5 \textcolor{blue}{(3)} & 78.3
      \\
      CGNet\cite{Yao2022} & 98.7 \textcolor{blue}{(2)} & 95.8 \textcolor{blue}{(1)} & 94.4 \textcolor{blue}{(1)} & 90.9 \textcolor{blue}{(2)} & 63.8 \textcolor{blue}{(5)} & 58.1 \textcolor{blue}{(5)} & 59.2 \textcolor{blue}{(6)} & 80.1
      \\
      Ours & 98.8 \textcolor{blue}{(1)} & 93.0 \textcolor{blue}{(4)} & 94.0 \textcolor{blue}{(2)} & 91.8 \textcolor{blue}{(1)} & 73.1 \textcolor{blue}{(3)} & 67.0 \textcolor{blue}{(2)} & 60.6 \textcolor{blue}{(4)} & \textbf{82.6}
      \\
\bottomrule
\end{tabular}
\end{table*}
 
\begin{table*}[ht]
    \centering
    \caption{ Comparative results of different methods on DsTok and SPL2018 at varying image resolutions. The highlighted values indicate the best results in each  case. The horizontal line indicates that the method did not produce any output at that resolution.}
    \label{tab:CG}
    \begin{tabular}{@{} *{10}{c}| *{9}{c} @{}}
      \Xhline{0.8pt}
      \multirow{3}*{Methods} & \multicolumn{9}{c}{DsTok}\vline &
      \multicolumn{9}{c}{SPL2018}\\
      \Xcline{2-10}{0.4pt} \Xcline{11-19}{0.4pt}
       & \multicolumn{3}{c}{224$\times$224} & \multicolumn{3}{c}{112$\times$112} & \multicolumn{3}{c}{56$\times$56}\vline  & \multicolumn{3}{c}{224$\times$224} & \multicolumn{3}{c}{112$\times$112} & \multicolumn{3}{c}{56$\times$56}\\
       \Xcline{2-10}{0.4pt} \Xcline{11-19}{0.4pt}
       & \scriptsize TPR & \scriptsize TNR & \scriptsize ACC & \scriptsize TPR & \scriptsize TNR & \scriptsize ACC & \scriptsize TPR & \scriptsize TNR & \scriptsize ACC & \scriptsize TPR & \scriptsize TNR & \scriptsize ACC & \scriptsize TPR & \scriptsize TNR & \scriptsize ACC & \scriptsize TPR & \scriptsize TNR & \scriptsize ACC\\
      \Xhline{0.4pt}
       ResNet18\cite{sha2022fake} & 77.1 & 73.9 & 75.5 & 65.4 & 71.4 & 68.4 & 62.2 & 57.3 & 59.7 & 85.3 & 85.3 & 85.3& 65.8 & 70.6 & 68.2 & 68.6 & 58.4 & 63.5 \\
       Quan\cite{quan2018} &  80.7 & 84.9 & 82.8 & 72.6 & 75.6 & 74.1 & -- & -- & -- & 89.5 & 89.6 & 89.6 & 85.9 & 83.2 & 84.6 & -- & -- & -- \\
      Yao\cite{yao2018}  &  86.5 & 91.6 & 89.1 & 87.8 & 84.2 & 86.0 & 79.6 & 74.5 & 77.1 & 90.1 & 87.4 & 88.7 & 89.4 & 84.1 & 86.8 & 88.0 & 79.9 & 83.9 \\
SPL2018\cite{SPL2018}    &  91.8 & 94.9 & 93.3 & 84.2 & 84.1 & 84.2 & 78.3 & 65.2 & 71.7  & 94.3 & 88.6 & 91.4  & 86.4 & 85.7 & 86.1 & 77.7 & 72.7 & 75.2\\
He\cite{He2020}         & 91.8 & 91.6 & 91.8 & 82.4 & 79.9 & 81.1 & 76.6 & 73.6 & 75.1 & 92.1 & 87.1  & 89.6 & 75.4 & 82.6 & 79.0 & 75.4 & 82.6 & 79.0\\
HcNet\cite{Zhang2020}      &  92.6 & 93.7 & 93.1 & 79.9 & 83.7 & 81.8 & 74.9 & 74.2 & 74.5 & 94.4 & 90.3 & 92.3 & 77.4 & 86.2 & 81.8 & 71.2 & 73.9 & 72.5\\    
QuanNet\cite{quan2020}       &  92.4 & 95.4 & 93.9 & 91.7 & 91.0 & 91.3 & 87.0 & 87.7 & 87.4 & \textbf{94.7} & 91.3 & 93.0 & 90.6 & 88.6 & 89.6 & 88.1 & 87.4 & 87.7\\   
CGNet\cite{Yao2022}      &  93.2 & 91.4 & 92.3 & 86.9 & 88.8 & 87.8 & 81.9 & 81.6 & 81.7 & 93.7 & 93.4 & 93.5 & \textbf{92.5} & 89.2 & 90.8 & 86.8 & \textbf{88.4} & 87.6 \\   
Ours       &  \textbf{98.1} & \textbf{96.9} & \textbf{97.5} & \textbf{95.3}& \textbf{93.6} & \textbf{94.4} & \textbf{92.8} & \textbf{90.4} & \textbf{91.6} & 94.4 & \textbf{94.2} & \textbf{94.3} & 92.2 & \textbf{90.1} & \textbf{91.2} & \textbf{89.2} & 87.6 & \textbf{88.4} \\ 
    \Xhline{0.8pt}
    \end{tabular}
\end{table*}

\begin{table*}[t!]
\centering
\caption{Ablation study about residual stream, content stream and  {Cross-Attention} module in the proposed model.  A checkmark  denotes the inclusion of the corresponding component in our model, while a cross  indicates its removal. The highlighted values denoted the best result in the corresponding case.}
\label{tab:ablation}
\begin{tabular}{*{3}{c}| *{12}{c}}
    \Xhline{0.8pt}
    \multicolumn{3}{c}{Ablation settings} & \multicolumn{3}{c}{DALL$\cdot$E2} & \multicolumn{3}{c}{DreamStudio} & \multicolumn{3}{c}{DsTok} & \multicolumn{3}{c}{SPL2018} \\
    \Xhline{0.4pt}
    Residual stream & Content stream & Cross-Attention &TPR & TNR & ACC & TPR & TNR & ACC & TPR & TNR & ACC & TPR & TNR & ACC \\
    \Xhline{0.4pt}
    $\checkmark$ & \ding{55} & \ding{55} & 99.1 & 99.0 & 99.0 & 98.9 & \textbf{99.6} & 99.3 & 93.2 & \textbf{98.6} & 95.9 & 92.2 & 93.3 & 92.8 \\
    \ding{55} & $\checkmark$ & \ding{55}  & 98.9 & 98.4 & 98.6 & 99.2 & 99.0 & 99.1  & 95.6 & 92.9 & 94.2 & 93.5 & 93.3 & 93.4 \\
    $\checkmark$ & $\checkmark$ & \ding{55}  & \textbf{99.3} & 98.8 & 99.0  & 99.4 & 99.4 & 99.4  & 95.9 & 96.9 & 96.4  & 92.3 & \textbf{94.2} & 93.2 \\
    $\checkmark$ & $\checkmark$ & $\checkmark$  & 99.2 & \textbf{99.2} & \textbf{99.2} & \textbf{99.5} & \textbf{99.6} & \textbf{99.5}  & \textbf{98.1} & 96.9 & \textbf{97.5}  & \textbf{94.4} & \textbf{94.2} & \textbf{94.3} \\
\Xhline{0.8pt}
\end{tabular}
\end{table*}

\subsection{Comparative Experiments} 
\label{subsec:comparative}
To highlight the efficacy of our model, we  {conduct} comparative studies against eight relevant techniques, including ResNet18 \cite{sha2022fake} and seven other contemporary methods for CG detection. These experiments are carried out on DALL$\cdot$E2 and DreamStudio across three distinct resolutions. The results are presented in Table \ref{Tab:main}. A review of these findings indicates that, in most instances, our proposed model outperforms the alternatives.
T2I poses a significant challenge when it comes to visual discernment. Despite this, our model, along with several CG detection algorithms such as CGNet\cite{Yao2022}, has demonstrated the ability to detect T2I with considerable effectiveness, especially when the image resolution is high. For instance, at the resolution of $256\times256$, the accuracies of several methodologies exceed 98.5\%. Remarkably, even when the resolution drops to $64\times64$, our model maintains an impressive performance, achieving 93.1\% and 97.8\% accuracy for the two T2I databases, respectively. 
It's worth noting that, although ResNet18 was advocated for T2I detection in the research \cite{sha2022fake}, our empirical results suggest that its detection performance falls short when compared to certain classic CG detection methods. 

\subsection{Robustness against Post-processing}
In this section, we assess the robustness of our proposed model in the face of seven post-processing techniques, encompassing adjustments to chromaticity, brightness, contrast, sharpness, rotation, and the application of Gaussian blur and mean blur. Chromaticity, brightness, contrast, and sharpness modifications are facilitated by the ImageEnhance function from the Pillow library, while the remaining alterations are executed using the OpenCV library. 
 To create a more realistic simulation of complex real-world scenarios, we've incorporated randomness into the parameters controlling the image alterations. For instance, the factors governing the degree of image manipulation (chromaticity, brightness, contrast) are randomly selected from a range of 0.5 to 2.5 for each image in the test dataset. Similarly, the factor controlling image sharpness is an arbitrary integer within the range of 0 to 4. Rotation degrees range from 0 to 360, and the kernel size for both Gaussian and mean filters is $5\times5$.
In our experimental procedures, we utilize the models obtained from the previous subsection \ref{subsec:comparative} to evaluate the post-processed test images directly. To keep the process simple, we restrict the detection accuracy display to image sizes of $256\times256$. The results of these experiments are illustrated in Table \ref{tab:post1} and Table \ref{tab:post2} for DALL·E2 and DreamStudio, respectively.\\
\indent Observations from these tables reveal that our proposed model exhibits superior performance in terms of average accuracy when compared to current methods. Additionally, we find that among all post-processing operations, current methods demonstrate comparatively lower robustness against rotation and blurring post-processing.

\subsection{CG Detection}
 In this section, we assess the performance of our model on two widely used datasets in the field of CG detection, namely SPL2018 \cite{SPL2018} and DsTok  \cite{DsTok}. The SPL2018 dataset consists of a comprehensive collection of 13,600 images. It includes both CG images obtained from over 50 game rendering software and professionally captured photorealistic PG images using various camera models under diverse environmental conditions. To create training, validation, and testing sets, we partitioned the dataset following a ratio of 10:3:4, respectively. The DsTok dataset comprises a total of 9,700 images sourced from the internet.  We partition them  into training, validation, and testing sets using a ratio of 3:1:1.  All test images are processed using center cropping with dimensions of $224\times224$, $112\times112$, and $56\times56$.  

We  conduct comparative experiments on both datasets at different resolutions, as presented in Table \ref{tab:CG}. The results demonstrate that our method consistently achieves the highest accuracy for both DsTok and SPL2018 at varying image resolutions. The improvement is particularly significant for DsTok. For example, our proposed method achieves accuracy improvements of 3.6\%, 3.1\%, and 4.2\% compared to the sub-optimal method QuanNet for the three resolutions. In the case of SPL2018, our accuracy outperforms CGNet by 0.8\%, 0.4\%, and 0.8\%, respectively.

\subsection{Ablation study}
 The proposed model, as depicted in Fig. \ref{fig:architecture}, incorporates two streams, namely the residual stream and the content stream, along with the cross-attention module  to enhance information exchange between the two streams. In this section, we  conduct  an ablation study to investigate the impact of the residual stream, content stream, and  cross-attention module  in our proposed model for both the T2I and CG detection tasks.  Among them, the residual stream and content stream are followed by the corresponding parts of feature fusion and classifier to achieve the task goal of binary classification.  The image resolutions used in our experiments are $256\times256$ and $224\times224$ for the respective tasks. The results of the ablation study are presented in Table \ref{tab:ablation}. From the results presented in Table \ref{tab:ablation}, we observe that employing all three components (i.e., both streams and the cross-attention module) in the proposed model consistently achieves the highest detection accuracy across all cases. When any of the components is removed, the performance of the corresponding model experiences varying degrees of decline in most cases. These  results strongly demonstrate the rationale behind the proposed model and reinforce its effectiveness for both the T2I and CG detection tasks.

\section{Conclusion}
In this study, we have directed our attention towards the Test-to-Image (T2I) generation process inherent in  AIGC, adopting an innovative, cross-attention enhanced dual-stream network for  T2I and  PG  detection. In an endeavor to appraise the effectiveness of our proposed method, we {generate} two distinct  T2I  graphics databases, leveraging the capabilities of DALL·E2 and DreamStudio systems. Through exhaustive comparative and ablation experiments, our method  affirms its superiority in detection performance, repeatedly surpassing traditional CG detection techniques across a diverse range of image resolutions. Importantly, our method  demonstrate  exceptional performance not just on our novel databases, but also excelled when applied to mainstream CG detection datasets, specifically DsTok and SPL2018.  
\par
This research heralds a substantial advancement in the field of AI-generated image forgery detection. Our innovative, cross-attention enhanced dual-stream network offers a promising solution to the escalating challenges introduced by AIGC. The effectiveness of our approach lays a compelling groundwork for further exploration and enhancement in this domain, with the potential to evolve into more sophisticated and dependable techniques for image forgery detection amid the dynamic environment of AI-generated content. Looking towards the future, we plan to further fine-tune this dual-stream network approach and broaden our investigations to its applicability in detecting other types of AI-generated forgeries, thus bolstering our comprehensive response to the expanding realm of artificial intelligence in content creation.

\printbibliography 

\end{document}